\documentclass[journal]{IEEEtran}

\ifCLASSINFOpdf
\else
\usepackage[dvips]{graphicx}
\fi
\usepackage{url}

\usepackage{graphicx}

\usepackage{epstopdf}
\usepackage{multirow}
\usepackage{array}
\usepackage{color}
\usepackage{lettrine}
\usepackage{pifont}
\usepackage{amsmath}
\usepackage{comment}
\usepackage{textcomp}
\usepackage{xcolor}
\DeclareMathOperator*{\argmin}{arg\,min}

\begin{document}

\title{IAIFNet: An Illumination-Aware Infrared and Visible Image Fusion Network}

\author{Qiao Yang, Yu Zhang, Zijing Zhao, Jian Zhang and Shunli Zhang

\thanks{This work was supported by the National Natural Science Foundation of China (No.61976017 and No. 62132002), the Beijing Natural Science Foundation (No.4202056) and the Fundamental Research Funds for the Central Universities (2022JBMC013) (\textit{Corresponding author: Jian Zhang})}
\thanks{Qiao Yang, Zijing Zhao, Jian Zhang, and Shunli Zhang are with the School of Software Engineering, Beijing Jiaotong University, Beijing 100044, China (e-mail: qyang\_bjtu@163.com; 21112202@bjtu.edu.cn; jianzh@bjtu.edu.cn; slzhang@bjtu.edu.cn).}
\thanks{Yu Zhang is with the School of Astronautics, Beihang University, Beijing 100083, China (e-mail: uzeful@163.com).}
}
\maketitle

\begin{abstract}
Infrared and visible image fusion~(IVIF) aims to create fused images that encompass the comprehensive features of both input images, thereby facilitating downstream vision tasks. However, existing methods often overlook illumination conditions in low-light environments, resulting in fused images where targets lack prominence. To address these shortcomings, we introduce the Illumination-Aware Infrared and Visible Image Fusion Network, abbreviated by IAIFNet. Within our framework, an illumination enhancement network initially estimates the incident illumination maps of input images, based on which the textural details of input images under low-light conditions are enhanced specifically. Subsequently, an image fusion network adeptly merges the salient features of illumination-enhanced infrared and visible images to produce a fusion image of superior visual quality. Our network incorporates a Salient Target Aware Module~(STAM) and an Adaptive Differential Fusion Module~(ADFM) to respectively enhance gradient and contrast with sensitivity to brightness. Extensive experimental results validate the superiority of our method over seven state-of-the-art approaches for fusing infrared and visible images on the public LLVIP dataset. Additionally, the lightweight design of our framework enables highly efficient fusion of infrared and visible images. Finally, evaluation results on the downstream multi-object detection task demonstrate the significant performance boost our method provides for detecting objects in low-light environments.
\end{abstract}


\IEEEpeerreviewmaketitle
\section{Introduction}
\label{sec:guidelines}

\IEEEPARstart{D}{ue} to the limited information captured by a single modality of the imaging device, the images obtained by multiple sensors are usually fused to produce a more comprehensive image about the monitored scene. In recent years, researchers have developed various methods to fuse multi-source images of the same scene as a high-quality image, which can strongly support the downstream computer vision tasks. 
In the field of image fusion, Infrared and Visible Image Fusion (IVIF) has been widely studied~\cite{LI2017100}. 
Specifically, infrared images capture and reflect the thermal radiation emitted from objects, making them suitable for detecting hidden targets. However, infrared images often lack significant textural details of the scene.
In contrast, visible images contain the majority of visual information within a monitored scene, making them highly suitable for perception by the human visual system~\cite{zhang2017infrared}.
Hence, the integration of the complementary features from the infrared image and visible image into a single fusion image is of utmost importance in fully comprehending the monitored scene.
Although IVIF has been widely studied and applied in the military domain~\cite{wong2017nitroaromatic, XIAO2023144}, semantic segmentation~\cite{tang2022image, TANG2023101870} and target detection~\cite{cao2019pedestrian,tang2023divfusion}, it continues to pose a significant challenge due to the low-quality images captured in low-light environments and the absence of ground-truth fusion images.
 
In recent decades, deep learning techniques have undergone significant development for IVIF. These techniques can be broadly categorized into three groups: auto-encoder (AE)-based \cite{li2018densefuse, li2021rfn, Reviewer2-3, 9502544, 9127964, 9349250}, convolutional neural network (CNN)-based \cite{liu2021smoa, ZHANG202099, xu2020u2fusion, di2022unsupervised, li2021different, liu2022coconet}, and generative adversarial network (GAN)-based methods \cite{ma2019fusiongan, ma2020ganmcc, 9894670, liu2022target, Reviewer2-1, Reviewer2-2}. DIDFuse~\cite{Reviewer2-3} pioneered the application of a decomposition network with a dual-stream structure to extract high and low-frequency features. SuperFusion~\cite{SuperFusion} integrates image registration, fusion, and the semantic requirements of high-level vision tasks into a comprehensive framework. Subsequently, SwinFusion~\cite{Reviewer1-1} and CDDFuse~\cite{Reviewer2-4} incorporate the Transformer structure, which models long-range dependencies, into the fusion network.
The above methods could well fuse images and facilitate downstream multi-object detection tasks in the daytime, but their performance would be significantly decreased in low-light environments.

\begin{figure*}
    \centering
    \includegraphics[width=1.98 \columnwidth, height=0.22\textwidth]{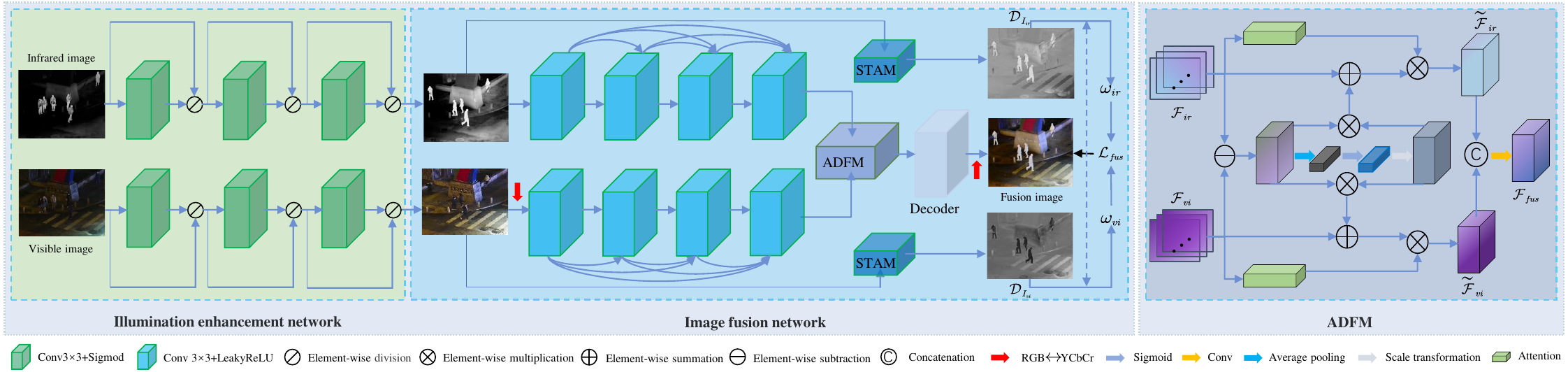}
    \vspace{-1mm}
    \caption{Flowchart of our method. The left block is the overall framework of our IAIFNet, which mainly consists of two modules, i.e., an illumination enhancement network and an image fusion network.
    The right block illustrates the structure of ADFM, and the bottom shows the symbols' meanings.}
    \label{figure1}
    \vspace{-4mm}
\end{figure*}

Recently, a few methods were proposed to restore details in dark regions of fused images. To obtain fused images with adaptive brightness, PIA~\cite{tang2022piafusion} balanced the brightness information from different source images by considering the illumination contribution. DIVFusion~\cite{tang2023divfusion} addressed the visual degradation defect of fused images in low-light environments by the Retinex theory~\cite{rahman2004retinex}. Although these works have achieved promising performance, there are still some drawbacks:
1) In the low-light environment, existing methods cannot efficiently balance the illumination characteristics of infrared and visible images to obtain high-quality fused images.
2) The difference between multiple modalities may be large in extreme environments, introducing lots of noise or color distortion in fused images.

To address the above challenges, in this work, we propose a novel infrared and visible image fusion framework based on illumination enhancement, namely IAIFNet. The main contributions of this work are mainly threefold:

\begin{itemize}
\item We propose a novel infrared and visible image fusion framework, which exploits illumination-aware information to effectively generate a fusion image of appropriate brightness and contrast.

\item We design two novel modules, i.e., an Adaptive Differential Fusion Module (ADFM) and a Salient Target Aware Module (STAM), in particular to solve the issues caused by the incident light, alleviating the influence of under-exposure or over-exposure and enhancing the salient targets in the fused images.

\item The proposed network is designed in a lightweight manner, thus it can quickly fuse infrared and visible images.
\end{itemize}

\section{The Proposed Method}
\label{sec:guidelines}

\subsection{Problem Formulation} 
The primary objective of IVIF is to create images that enhance both visual inspection and computer perception in low-light environments. 
To address this, the Retinex model is commonly employed to decompose the observed image. However, in low-light environments, noise in this decomposition process cannot be overlooked. 
Therefore, in this paper, the model is defined as follows:
\begin{equation}
    I=L\odot \left( R+N \right) = L\odot I^{en}
\end{equation}
where $L$, $R$ and $N$ represent the incident illumination map, the reflection map and the noise map, respectively. Further, we formulate the illumination-oriented fusion as an optimization model:
\begin{equation}
\min_{\omega _u}\ \mathcal{L}_{fus}\left( I_{fus},\psi \left( I_{ir}^{en,*},I_{vi}^{en,*};\omega _u \right) \right) 
\end{equation}
\begin{equation}
s.t.\ \left\{ \begin{array}{l}
	I_{ir}^{en,*}\in \underset{I_{ir}^{en}}{\argmin}\ \mathcal{L}_{ill}\left( I_{ir}^{en},\phi \left( I_{ir} \right) ;\omega _i \right)\\
	I_{vi}^{en,*}\in \underset{I_{vi}^{en}}{\argmin}\ \mathcal{L}_{ill}\left( I_{vi}^{en},\phi \left( I_{vi} \right) ;\omega _i \right)\\
\end{array} \right. 
\end{equation}
where $\mathcal{L}_{fus}$ denotes the fusion-specific loss between illuminance-based fused image $I_{fus}$, the enhanced image $I_{ir}^{en}$, $I_{vi}^{en}$. This further divides the problem into two subproblems separately, to develop the fusion model $\psi$ based on enhanced images, and to achieve the enhanced images based on the constraints.
It is necessary to construct illumination enhancement network $\phi$ for infrared and visible images, respectively. Additionally, an enhance-specific loss $\mathcal{L}_{ill}$ will help to evaluate the performance of these enhancement networks.
The learnable parameters in the fusion network $\psi$ and the enhancement network $\phi$ are denoted as $\omega _u$ and $\omega _i$, respectively. 
In subsequent sections, we will go through the specific designs of $\psi$ and $\phi$, as well as the precise definition of $\mathcal{L}_{fus}$.

\subsection{Overall Architecture} Fig.~\ref{figure1} illustrates the overall architecture of our image fusion framework with two modules. In the illumination enhancement network, the sequential convolutional layers with sigmoid activation function extract the illumination-aware features from infrared and visible images through a residual connection based on Retinex theory. 
According to the generated distinct illumination-related features, the enhanced images can be further obtained. 
In the image fusion network, the sequential convolutional layers with leakyReLU activation function use the skip connection to extract multiple levels of feature maps, while ADFM is designed to integrate complementary and differential information in these feature maps. Then STAM captures the salient target area from the enhanced images. 
Inspired by \cite{DIP}, \cite{RetinexDIP}, \cite{SDNet}, image fusion network is constructed as an encoder-decoder consisting of convolution blocks. Finally, the integrated feature is fed into the subsequent decoder to reconstruct the fusion image. 
Consequently, IAIFNet is a lightweight network with a small number of parameters, enabling quickly inference speed.

\subsection{Network Details}
\textbf{Illumination Enhancement Network.}{ As the infrared and visible images are simultaneously taken at the same place, their illumination conditions can be regarded as the same. Inspired by~\cite{ma2022toward}, we leverage a single network to estimate the illumination map for both modalities of images, quickly enhancing the brightness of low-light images by sharing parameters. The image enhancement process can be expressed as 
\begin{equation} 
\setlength{\abovedisplayskip}{3pt}
I_{ir}^{en}=I_{ir}\oslash{L_{ir}} ,
I_{vi}^{en}=I_{vi}\oslash{L_{vi}} ,
\setlength{\belowdisplayskip}{3pt}
\end{equation}
where ${L_{ir}}$ and ${L_{vi}}$ denote the estimated illumination maps of infrared image $I_{ir}$ and visible image $I_{vi}$, respectively. $I_{ir}^{en}$ and $I_{vi}^{en}$ denote the enhanced images. 
$\oslash$ denotes the element-wise division operator. 
Besides, in this study, we estimate the illumination map in the RGB color space to maintain consistency with color information.

\textbf{Image Fusion Network.}{ In order to obtain a fusion image with more complementary features and better exposure status, we design an image fusion network with two novel modules, i.e., ADFM and STAM.}
\subsubsection{Network Architecture} Inspired by \cite{li2018densefuse} and \cite{xu2020u2fusion}, we first design a feature extraction block with skip connections to make use of the features in multiple levels for comprehensive representation. In the feature fusion stage, we design the ADFM with the attention mechanism \cite{di2022unsupervised} and the differential operation \cite{tang2022piafusion} in order to adjust brightness and reduce redundant information (please see Sec.\ref{exp} for detailed discussions). Specifically, to adaptively stress the importance of the features, the attention mechanism is introduced in ADFM and the attention weight map is constructed as
\begin{equation} 
\mathcal{A}_{tt}=\mathcal{S}\left( Conv_{3\times3}\left( \mathcal{F}_{ir} \right) \otimes Conv_{3\times3}\left( \mathcal{F}_{vi} \right) \right) .
\setlength{\abovedisplayskip}{6pt}
\end{equation}
where $\mathcal{F}_{ir}$ and $\mathcal{F}_{vi}$ represent the extracted multi-level feature maps of infrared and visible images, respectively. 
Further, we exploit $\mathcal{A}_{tt}$ and difference map of $\mathcal{F}_{vi}$ and $\mathcal{F}_{ir}$ to finely tune $\mathcal{F}_{vi}$ and $\mathcal{F}_{ir}$ as:
\begin{equation} 
\widetilde\mathcal{F}_{ir}=\left( \mathcal{S}\left( \mathcal{G}\left( \mathcal{F}_{vi}\ominus \mathcal{F}_{ir} \right) \right) \odot \left( \mathcal{F}_{vi}\ominus \mathcal{F}_{ir} \right) \oplus \mathcal{F}_{ir} \right) \otimes \mathcal{A}_{tt} ,
\end{equation}
\begin{equation} 
\setlength{\abovedisplayskip}{6pt}
\widetilde\mathcal{F}_{vi}=\left( S\left( \mathcal{G}\left( \mathcal{F}_{ir}\ominus \mathcal{F}_{vi} \right) \right) \odot \left( \mathcal{F}_{ir}\ominus \mathcal{F}_{vi} \right) \oplus \mathcal{F}_{vi} \right) \otimes
\setlength{\belowdisplayskip}{6pt}\mathcal{A}_{tt} ,
\end{equation}
where $\widetilde\mathcal{F}_{ir}$ and $\widetilde\mathcal{F}_{vi}$ represent corrected feature. $\mathcal{S}\left( \cdot \right)$, $\mathcal{G}\left( \cdot \right)$ and $\odot$ denote sigmoid function, global average pooling and channel-wise multiplication, respectively. Then, the fused feature map $\mathcal{F}_{fus}$ is obtained by 
\begin{equation} 
\setlength{\abovedisplayskip}{6pt}
\mathcal{F}_{fus}=\mathcal{C}onv_{3\times3}\left(\mathcal{C}at\left( \widetilde\mathcal{F}_{ir}, \widetilde\mathcal{F}_{vi} \right) \right) ,
\setlength{\belowdisplayskip}{6pt}
\end{equation}
where $\mathcal{C}at\left( \cdot \right)$ represents channel concatenation operation. Finally, the fusion image $I_{fus}$ with different modalities of salient features is reconstructed from $\mathcal{F}_{fus}$ by the decoder. 
The entire fusion process is implemented in YCbCr space, ensuring that the color channel information is kept consistent with the fused brightness information, thereby reducing color bias.

\begin{figure*}[!t]
\centering
\centerline{\includegraphics[width=1.95 \columnwidth, height=0.18\textwidth]{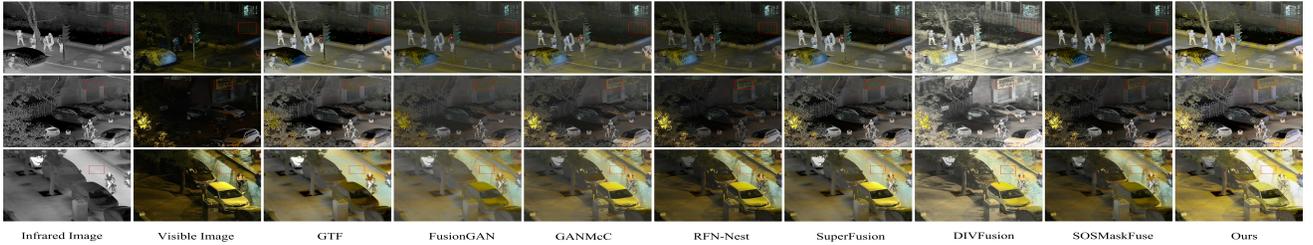}}
\vspace{-2mm}
\caption{Qualitative comparison of our IAIFNet and seven state-of-the-art methods on fusing three pairs of infrared and visible images (i.e., \#060193, \#120089 and \#190001)). In each image, one red region is annotated for clear comparison.}
\vspace{-3mm}
\label{figure2}
\end{figure*}

\subsubsection{Loss Function}{ The loss function of our image fusion model can be defined as
\begin{equation} 
\setlength{\abovedisplayskip}{3pt}
\mathcal{L}_{fus}=\alpha \mathcal{L}_{struct}+\beta \mathcal{L}_{smooth}^{int}+\gamma \mathcal{L}_{grad} ,
\setlength{\belowdisplayskip}{3pt}
\end{equation}
where $\mathcal{L}_{struct}$, $\mathcal{L}_{smooth}^{int}$, and $\mathcal{L}_{grad}$ denote the structure loss, intensity consistent loss and gradient loss, 
respectively, and $\alpha$, $\beta$, and $\gamma$ are the corresponding hyperparameters to adjust their weights. 
Inspired by \cite{liu2021smoa, liu2022target, ma2017infrared}, in order to fully integrate the salient features of the infrared and visible images, we develop STAM to compute the salience map of each image $I$, where the saliency value $\mathcal{D}_I\left( x \right)$ at pixel $x$ is computed as
\begin{equation} 
\mathcal{D}_I\left( x \right) = \mathcal{H}_I \mathcal{M}_p^\top ,
\end{equation}
where $\mathcal{M}_p=[l_p^0,l_p^1,\cdots,l_p^{255}]\in {R}^{1\times 256}$ and $l_p^i=(x-i)^p$, $p$ is used to adjust the degree of objects' saliency. $\mathcal{H}_I\in {R}^{1\times 256}$ is the histogram of $I$. To be specific, $\mathcal{L}_{struct}$ is formulated as
\begin{equation} 
\mathcal{L}_{struct}=1-SSIM\left( I_{fus}\ ,\ \omega _{ir}\otimes I_{ir}^{en}+\omega _{vi}\otimes I_{vi}^{en} \right) ,
\end{equation}
where $\omega _{ir}=0.5+\left( \mathcal{D}_{I_{ir}^{en}}-\mathcal{D}_{I_{vi}^{en}} \right) /2,\ \omega _{vi}=1-\omega _{ir}$, $\mathcal{D}_{I_{ir}^{en}}$ and $\mathcal{D}_{I_{vi}^{en}}$ represent saliency maps of enhanced infrared and visible images, respectively.
$SSIM(A,B)$~\cite{wang2004image} calculates the structural similarity of images $A$ and $B$. Thus, $\mathcal{L}_{struct}$ can supervise the network to transfer the diverse structural details from input images to the fusion image. Then, $\mathcal{L}_{smooth}^{int}$ is formulated as
\begin{equation} 
\mathcal{L}_{smooth}^{int}={\Vert } I_{fus}-\left(\omega _{ir}\otimes I_{ir}^{en}+\omega _{vi}\otimes I_{vi}^{en} \right){\Vert _1} ,
\end{equation}
where ${\Vert } \cdot {\Vert _1}$ denotes the $l_1$-norm. In this work for the fused images, $\mathcal{L}_{smooth}^{int}$ not only constrains the pixel-level intensity distribution but also assists our illumination enhancement network in suppressing the illumination information smoothly \cite{wei2018deep}. Finally, inspired by~\cite{SDNet}, $\mathcal{L}_{grad}$ was designed to reduce the effect of noise in the fusion process by incorporating the fusion network, and it formulated as
\begin{equation} 
\mathcal{L}_{grad}=\Vert \nabla I_{fus}-max\left( \nabla I_{ir}^{en},\nabla I_{vi}^{en} \right) \Vert _1 ,
\end{equation}
where $\nabla$ and $max\left(\cdot\right)$ denote the Sobel gradient operator and elementwise-maximum selection operator, respectively. 
This gradient loss can supervise our model to preserve more salient edges and textures from the source images into the fusion image.
In all experiments of this work, $p$ is set to 2, and $\alpha$, $\beta$, and $\gamma$ are set to 1, 15, and 3, respectively.

\begin{figure}
\centering
    \includegraphics[width=0.95 \columnwidth, height=0.185\textwidth]{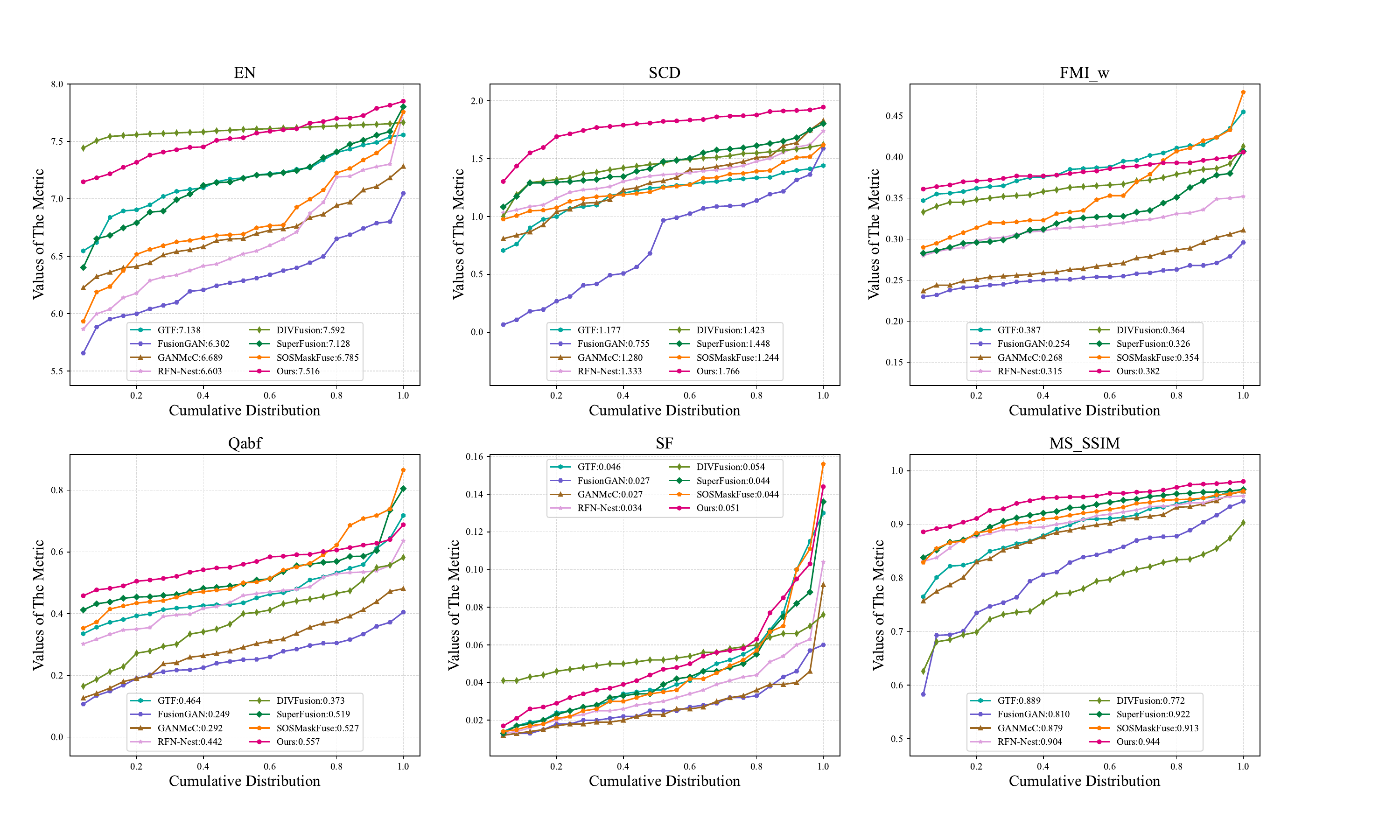}
    \vspace{-2mm}
    \caption{Quantitative comparisons of different image fusion methods. A single point ($x,y$) on the curve denotes that there are ($100\times x$) \% percent of image pairs that have metric values no more than y.}
    \label{figure3}
\end{figure}

\section{Experimental Results and Discussions}\label{exp}
\label{sec:guidelines}

\subsection{Dataset and Implementation Details}
LLVIP is a famous infrared and visible image dataset captured at night and has been widely used for evaluating the performance of image fusion and object detection methods~\cite {jia2021llvip}. In this study, 240 pairs of infrared and visible images from LLVIP are used for training and 50 pairs for the test. The whole network is implemented based on PyTorch, trained on an NVIDIA RTX 3090 GPU, and the size of input image patches is set to 600$\times$400. 
During the training of the illumination enhancement network, the batch size is set to 8, and the epoch number is set to 100. When training the image fusion network, the illumination-enhanced infrared and visible images are taken as its inputs, the batch size is set to 6, the epoch number is set to 150, and the initial learning rate is set to 0.001 and then decayed by 10 every 30 epochs. 
The Adam~\cite{kingma2014adam} optimizer ( $\beta_1$=0.9 and $\beta_2$=0.999 ) is exploited to optimize the parameters of our complete image fusion network. 

\subsection{Performance Analysis}
We conduct experiments against a series of state-of-the-art (SOTA) methods, including GTF~\cite{ma2016infrared}, FusionGAN~\cite{ma2019fusiongan}, GANMcC~\cite{ma2020ganmcc}, RFN-Nest~\cite{li2021rfn}, SuperFusion~\cite{SuperFusion}, DIVFuison~\cite{tang2023divfusion}, and SOSMaskFuse~\cite{SOSMaskFuse} to verify the effectiveness of our IAIFNet.

\textbf{Qualitative Comparisons.} Three comparison examples from the LLVIP dataset (\#060193, \#120089, and \#190001) are depicted in Fig. 2. Our IAIFNet demonstrates two significant advantages over other methods. Firstly, it adeptly preserves salient targets from both source images, as highlighted by regions in red boxes in the first and third rows of Fig. 2, ensuring enhanced contrast for visual observation. Secondly, our results effectively retain color information from visible images, as evident from the regions in red boxes in the second row, aligning closely with the human visual system. In contrast, methods such as GTF, FusionGAN, GANMcC, and RFN-Nest, which overlook the illumination condition of source images, yield results with inferior visual quality, failing to adequately preserve contrast and details. Both DIVFusion and our IAIFNet address this illumination issue by leveraging the Retinex theory. However, DIVFusion results suffer from overexposure and noise defects.

\begin{table}[t]
\caption{The Overall Model Size and Inference Time \\
for Processing 50 Pairs of Images with Size 256$\times$256.}
\label{tab2}
\resizebox{\linewidth}{!}{
\begin{tabular}{ccccccccc}
\hline
Method  & GTF & FusionGan & GANMcC & RFN-Nest &SuperFusion & DIVFusion & SOSMaskFuse & Ours  \\ \hline
Params(M)$\downarrow$   & -   & \underline{1.982}     & 2.271  & 30.097 &2.101  & 4.403  &36.951   & \textbf{0.635} \\ \hline
Time(s)$\downarrow$ & -   & 0.677     & 1.295  & 0.246    &\underline{0.023} & 0.087   &0.087  & \textbf{0.011} \\ \hline
\end{tabular}}
\vspace{-3mm}
\end{table}

\begin{table}[t]
\centering
\caption{Multi-Object Detection Results on the LLVIP Dataset.}
\label{tab1}
\resizebox{\linewidth}{!}{
\begin{tabular}{ccccccc}
\hline
\multirow{2}{*}{} & \multirow{2}{*}{Precision$\uparrow$} & \multirow{2}{*}{Recall$\uparrow$} & \multicolumn{3}{c}{mAP@.5$\uparrow$} & \multirow{2}{*}{mAP@{[}.5:.95{]}$\uparrow$} \\ \cline{4-6}
 &  &  & Person & Car & Mean &  \\ \hline
Infrared & 0.834 & 0.476 & 0.797 & 0.556 & 0.677 & 0.476 \\ \hline
Visible & 0.833 & 0.654 & 0.836 & 0.713 & 0.775 & 0.562 \\ \hline
GTF & 0.873 & 0.635 & 0.829 & 0.725 & 0.777 & 0.569 \\ \hline
FusionGAN & 0.859 & 0.585 & 0.815 & 0.678 & 0.747 & 0.552 \\ \hline
GANMcC & 0.858 & 0.695 & 0.837 & 0.793 & 0.815 & 0.621 \\ \hline
RFN-Nest & 0.890 & \textbf{0.726} & 0.837 & \textbf{0.830} & \underline{0.834} & \underline{0.631} \\ \hline
SuperFusion & \textbf{0.917} & 0.601 & 0.829 & 0.725 & 0.778 & 0.587 \\ \hline
DIVFusion & 0.835 & 0.664 & 0.759 & 0.811 & 0.785 & 0.598 \\ \hline
SOSMaskFuse & 0.853 & 0.704 & \textbf{0.860} & 0.752 & 0.806 & 0.613 \\ \hline
Ours & \underline{0.907} & \underline{0.705} & \underline{0.854} & \underline{0.823} & \textbf{0.839} & \textbf{0.633} \\ \hline
\end{tabular}}
\vspace{-3mm}
\end{table}

\textbf{Quantitative Comparisons.}
As shown in Fig.~\ref{figure3}, we use six metrics to compare quality of the fusion images produced by different methods, including entropy (EN), sum of correlation of difference (SCD) \cite{aslantas2015new}, feature mutual information with wavelet transform (FMI\_w) \cite{haghighat2014fast}, edge information transfer (Q$_{abf}$) \cite{xydeas2000objective}, spatial frequency (SF)~\cite{SF} and multi-scale structural similarity index measure~(MS\_SSIM)~\cite{ma2015perceptual}.
The greatest values of SCD, Q$_{abf}$, and MS\_SSIM indicate that our fusion images preserve more structural information and higher contrast from the source images. Additionally, our method obtains the second-best results in terms of EN, SF and FMI\_w.

Further, we evaluate the memory consumption and computation efficiency of the comparison methods. 
Specifically, the average inference time is computed for input images of size 256$\times$256. 
As presented in Table \ref{tab2}, our model performs best in terms of both model size and inference speed. 

\begin{figure}[t]
\centering
\centerline{\includegraphics[width=1.0 \columnwidth]{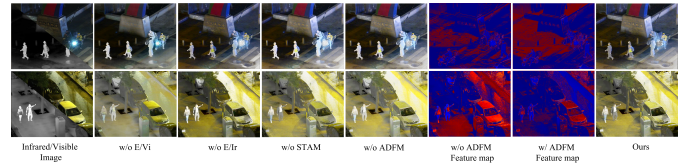}}
\vspace{-3mm}
\caption{Qualitative comparisons of different strategies on \#190271 and \#010045, respectively.}
\label{figure5}
\end{figure}

\subsection{Performance on Object Detection Task} To further explore the influence of our method on high-level vision tasks, we apply it to the multi-object detection task. Specifically, we select YOLOv5 \cite{redmon2016you} as the baseline model to detect pedestrians and vehicles from the fusion images. As shown in Table \ref{tab1}, our method outperforms other methods on multi-object detection. In terms of mAP@[.5:.95] on LLVIP, the detection method with our fusion images gained 7.1\% improvement over that using only visible images, and outperformed those with fusion images of other methods. 

\begin{table}[!t]
\caption{Quantitative Comparisons of Different Strategies.}
\label{tab3}
\resizebox{\linewidth}{!}{
\begin{tabular}{cccccccccc}
\hline
E/Vis & E/Inf & STAM & ADFM & \multicolumn{1}{c}{EN$\uparrow$} & \multicolumn{1}{c}{SCD$\uparrow$} & \multicolumn{1}{c}{FMI\_w$\uparrow$} & \multicolumn{1}{c}{Q$_{abf}$$\uparrow$} & SF$\uparrow$ & MS\_SSIM$\uparrow$ \\ \hline
\ding{55} & \ding{51} & \ding{51} & \ding{51} & 7.385 & 1.482 & 0.354 & 0.414 & 0.039 & 0.863 \\ \hline
\ding{51} & \ding{55} & \ding{51} & \ding{51} & 7.157 & 1.655 & 0.338 & \textbf{0.578} & \textbf{0.059} & \underline{0.943} \\ \hline
\ding{51} & \ding{51} & \ding{55} & \ding{51} & 7.298 & 1.670 & \textbf{0.401} & 0.553 & 0.043 & 0.935 \\ \hline
\ding{51} & \ding{51} & \ding{51} & \ding{55} & \underline{7.501} & \textbf{1.771} & {0.369} & 0.547 & \underline{0.056} & \underline{0.943} \\ \hline
\ding{51} & \ding{51} & \ding{51} & \ding{51} & \textbf{7.516} & \underline{1.766} & \underline{0.382} & \underline{0.557} & 0.051 & \textbf{0.944} \\ \hline
\end{tabular}}
\vspace{-4mm}
\end{table}

\subsection{Ablation Analysis}
We design ablation experiments to test the effectiveness of the illumination enhancement network, STAM and ADFM, as shown in Table~\ref{tab3}, where ``E/" indicates the illumination enhancement for corresponding image.
The results indicate that the fusion image by enhancing the illumination of both infrared and visible images is much better than those by only enhancing a single image, and STAM can further strengthen the performance of our IAIFNet effectively. Fig.~\ref{figure5} demonstrates that the fusion image without using ADFM suffers the under-exposure or over-exposure defect, as indicated by the feature maps in columns six and seven. 
It can be seen that our fusion images demonstrate the best visual effect in terms of illumination distribution and detail visualization, indicating that the proposed ADFM can adaptively adjust the brightness of the fusion image and highlight its important details.

\section{Conclusion}
In this letter, we propose a novel illumination-aware infrared and visible image fusion method, which outperforms other SOTA methods especially on fusing images captured in low-light environments. 
Moreover, our method can not only improve the visual quality of the fusion image effectively through illumination enhancement, but also boost the performance of downstream multi-object detection task significantly.

\clearpage

\bibliographystyle{IEEEtran}
\IEEEtriggeratref{24}
\bibliography{ref}

\end{document}